\documentclass[letterpaper, 10 pt, conference]{ieeeconf}
\IEEEoverridecommandlockouts
\usepackage{graphicx}
\usepackage[hidelinks]{hyperref}
\let\labelindent\relax
\usepackage{enumitem}
\graphicspath{{figures/}{figures/final/}{figures/}}
\usepackage[T1]{fontenc}   %

\usepackage{amsmath, amssymb, amsfonts, amsthm}
\usepackage[dvipsnames]{xcolor}
\usepackage{array, tabularx, booktabs, multirow, makecell}
\usepackage[ruled,linesnumbered,lined]{algorithm2e}
\usepackage[font=footnotesize,labelsep=period,skip=5pt]{caption}

\theoremstyle{remark}

\definecolor{MyPurple}{HTML}{800080}
\definecolor{MyOrange}{HTML}{d38100}
\definecolor{MyGreen}{HTML}{00bf20}
\definecolor{MyBlue}{HTML}{1455a6}

\usepackage{lipsum}

\title{
\LARGE \bf 
CONSORT: Uncertainty-Constrained Multi-Robot Planning for Cooperative Localization in GPS-Denied Environments
}

\author{Tony Zhang$^{1,2}$, Ryan Bartholomew$^{1}$, Kalliyan Velasco$^{1}$, Joshua Mangelson$^{1}$, Brady Moon$^{1}$%
\thanks{This work was supported by the NSF-funded BYU Robotics Pioneers REU program (Award No.~2447307) and in part by the NAVSEA NEEC (Award No.~N00174-23-1-0005) in coordination with NSWC PCD.}
\thanks{$^{1}$Authors are with the FROST Lab, Ira A. Fulton College of Engineering, Brigham Young University, Provo, UT 84602, USA.}%
\thanks{$^{2}$T. Zhang is with the Kahlert School of Computing, University of Utah, Salt Lake City, UT 84112, USA, and contributed to this work as a visiting undergraduate researcher in the FROST Lab.}
}

\begin{document}

\maketitle
\thispagestyle{empty}
\pagestyle{empty}

\begin{abstract}
Centralized multi-agent path planning with cooperative localization in GPS-denied environments is generally computationally 
intractable when formulated in full belief space due to tightly coupled uncertainty dynamics. Prior work typically approximates belief dynamics
or restricts inter-agent coupling. This work examines a multi-agent planning problem in which a primary agent must reach a goal in minimum time 
subject to a chance-constrained collision-avoidance requirement and a terminal bound on localization uncertainty. We obtain a tractable surrogate
of the underlying belief-space problem by optimizing over the mean-state trajectories of all agents and propagating covariance for terminal constraint evaluation only. 
We propose CONSORT (Cooperative Navigation with Support-Optimized Routing and Trajectories), a discrete-to-continuous pipeline combining hexagonal-grid joint search with spline-based trajectory refinement. 
Across 50 randomized scenarios, the planners separate as the terminal uncertainty bound tightens, with CONSORT solving 80\% of scenarios where the strongest baseline solves 54\%, at comparable path length. 
In marine simulations, with the plans executed under feedback control on an independently developed factor-graph estimate, the  trajectories remain collision-free and their terminal uncertainty falls within 6\% of the planner's prediction.

\end{abstract}
\section{Introduction}

In environments where GPS signal is unreliable, such as underwater, or in regions subject to spoofing or jamming, agents must rely on other forms of information
for localization. In these environments, autonomous agents struggle with state uncertainty due to inertial measurement unit (IMU)
drift. As the estimation error accumulates, an agent's pose uncertainty can grow to a point where waypoint tracking degrades and mission objectives are compromised.
Common planning strategies ameliorate this by biasing trajectories toward information-rich regions~\cite{roy_coastal_1999} and introducing agents to utilize inter-agent measurements for 
cooperative localization~\cite{roumeliotis2002dist,bahr2009coop}. Consequently, trajectory planning must account for uncertainty propagation when estimation error impedes task performance. 

As the joint belief of a team grows exponentially with its size and inter-agent measurements couple the agents' beliefs, we optimize mean trajectories and propagate uncertainty only to evaluate constraints. 

\begin{figure}[t]
  \centering
  \includegraphics[width=\columnwidth]{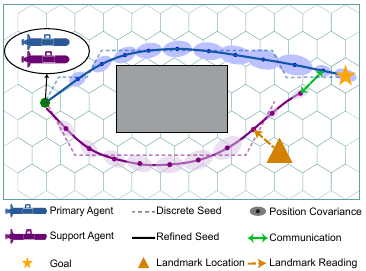}
  \caption{The escort problem. The primary must reach the goal with its
    localization uncertainty under a terminal bound, and the support is routed
    to hold it there. \textcolor{MyPurple}{Support agent} \textcolor{MyOrange}{localizes with landmark} and \textcolor{MyGreen}{relays~position~data}, enabling the \textcolor{MyBlue}{primary agent} to take a more direct path. Both trajectories are searched on the heading-aware
    hexagonal lattice and refined as B-splines.}
  \label{fig:overview}
  \vspace{-1.5em}  
\end{figure}

In this work, we consider a centralized multi-agent path planning with cooperative localization (MAPP-CL) problem, a variant of the CL-MRMP problem \cite{theurkauf2025coop}, in which a primary agent must reach a goal in minimum time while subject to a chance-constrained collision-avoidance 
requirement and a terminal bound on localization uncertainty. We call this the escort problem: the support agents have no goal locations and are routed solely to hold the primary's terminal uncertainty under the constraint (Fig.~\ref{fig:overview}).

We propose CONSORT (Cooperative Navigation with Support-Optimized Routing and Trajectories), a centralized planner for this escort problem.\footnote{The Julia implementation, 50 benchmark scenarios, and simulation videos are available on the project page \href{https://frostlab.byu.edu/consort-planner/}{frostlab.byu.edu/consort-planner/}.}
We contribute the following:
\begin{enumerate}[nosep, leftmargin=*]
  \item A MAPP-CL formulation that jointly routes support and primary agents under a terminal uncertainty bound and chance-constrained obstacle avoidance.
  \item CONSORT, a planner pairing joint heading-aware hexagonal lattice search with B-spline refinement, both enforcing curvature, collision, and uncertainty constraints.
  \item A benchmark of 50 randomized scenarios on which CONSORT solves 80\% at the 50\% constraint level, against 54\% for the strongest of four baselines.
  \item Monte Carlo validation in HoloOcean~\cite{potokar2024holoocean} with an independent factor-graph estimator where plans stay collision-free and terminal uncertainty is within 6\% of prediction.
\end{enumerate}

\section{Related Work}
\label{sec:related}
Planning under uncertainty is naturally posed as a partially observable Markov
decision process \cite{kaelbling1998pomdp}, which is intractable in general
\cite{papadimitriou1987mdp}.
Belief-space planners restrict beliefs to Gaussians and propagate them along
candidate trajectories, over a roadmap \cite{prentice2009brm}, a tree
\cite{bry2011rrbt}, or by ranking candidate paths under a closed-loop model \cite{vandenberg2011lqgmp},
usually under the assumption that the most likely observation is received
\cite{platt2010belief}. These methods plan for a single agent, and the
multi-robot case of LQG-MP plans the robots one at a time without
inter-agent measurements. With several
agents the joint belief and the search space grow exponentially with the
size of the team.

\begin{figure*}[t]
  \centering
  \includegraphics[width=\textwidth]{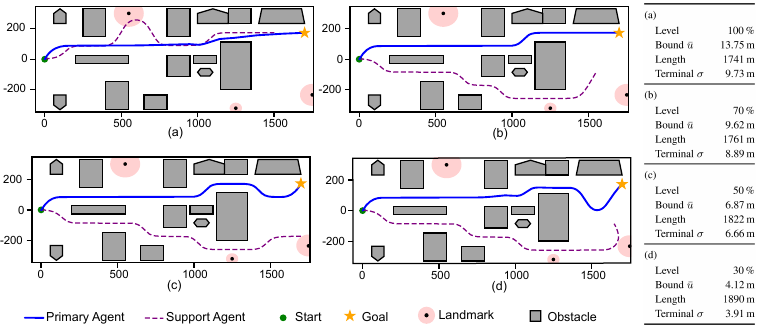}
  \caption{Constraint ladder on one scenario: a 1700\,m corridor of staggered
    walls, 14 obstacles and three landmarks. The terminal bound tightens from
    (a) to (d), and every tightening is paid for in primary length; the levels
    and the solutions they produce are listed at right. In (a) the
    support climbs into the north pocket for the weak landmark there and
    rejoins; in (b) it takes the southern lane to a stronger landmark and
    relays it; in (c) the primary adds a turn before
    the goal so that the support can reach the second landmark; in (d) it dips to the
    center line to stall allowing the support to relay from a closer distance.}
    \vspace{-0.3cm}
  \label{fig:ladder}
\end{figure*}

Cooperative localization lets agents share pose estimates and measure one
another to reduce dead-reckoning drift \cite{roumeliotis2002dist,bahr2009coop}
and is established in underwater navigation \cite{paull2014auv}. Because cooperating agents share
measurement history, naive fusion double counts information, and covariance
intersection \cite{julier1997cov} is the usual remedy \cite{carrillo2013ci}.
Planning with cooperative localization in the loop is more recent. Theurkauf
et al. \cite{theurkauf2025coop} grow a sampling-based tree over the joint
belief of a team, and Liu et al. \cite{liu2026risk} plan risk-aware
trajectories for underwater teams. Our support agents have no goals of their
own and are routed only to meet the primary's terminal bound, and our joint
search is exhaustive on a lattice rather than sampled.

Multi-agent path finding solves the discrete coordination problem exactly
with conflict-based search \cite{sharon2015cbs} or approximately by planning
the agents one at a time in priority order
\cite{erdmann1987multiple,silver2005cooperative}, but it treats positions as
known. Hexagonal grids give six equidistant neighbors and paths that are
easier to smooth \cite{duszak2021hexgrid}, and B-splines give smooth
trajectories whose curvature can be bounded \cite{feng2024bsrrt}. Obstacle
avoidance under position uncertainty extends the chance-constrained
formulation of Blackmore et al. \cite{blackmore2011chance} to uncertain
obstacle positions, and we bound each spline segment by its MINVO hull
\cite{tordesillas2022minvo} so that the constraint is linear in the control
points.

\section{Problem Formulation}
\label{sec:pf}
Let $\mathcal{I} = \{1,2,\dots,P\}$ denote a team of $P$ agents, where agent $i=1$
is the primary agent tasked with reaching a goal location $G\in\mathbb{R}^2$.
All other agents $2,\dots,P$ are support agents whose sole task is to assist the
primary agent's localization. All agents move at a common constant speed $v$,
so minimizing time is equivalent to minimizing trajectory length, and we use
arc length $s = vt$ as a shared clock throughout. The environment contains $M$
static convex obstacles $\mathcal{O}_1,\dots,\mathcal{O}_M$, each defined as an
intersection of half-planes $a_j^{\top}x \le b_j$ with unit outward normals
$a_j$ and offsets $b_j$, and each located only up to a covariance $\Sigma_o$.
The environment also contains a set $\mathcal{L}$ of landmarks with known
positions $\ell\in\mathbb{R}^2$ and positional uncertainty $\Sigma_\ell$.

\subsection{Decision Variables}

We optimize over mean-state trajectories only. The decision variables are the
heading profile $\theta_i(s)$ of every agent and the primary's path length
$S_1$. Positions follow from the headings at the arc samples
$s_k = k\,\Delta s$ through
\begin{equation}
  x_i(s_{k+1}) = x_i(s_k) + \Delta s
  \begin{bmatrix} \cos\theta_i(s_k) \\ \sin\theta_i(s_k) \end{bmatrix}.
  \label{eq:dyn}
\end{equation}

\subsection{Covariance Propagation}

Let $\Sigma_i(s)$ denote the $2\times2$ position covariance of agent $i$.
Covariance is not a decision variable. It is propagated along the mean
trajectories and used only to evaluate constraints. Between measurements the
agents dead-reckon, and we model the drift as process noise whose variance
grows linearly with distance and is larger along the direction of travel than
across it:
\begin{equation}
  \Sigma_i^{-}(s_{k+1}) = \Sigma_i(s_k) + Q\big(\theta_i(s_k)\big)\,\Delta s,
  \label{eq:prop}
\end{equation}
where $Q(\theta) = R(\theta)\,\mathrm{diag}(q_\parallel, q_\perp)\,R(\theta)^{\!\top}$,
$R(\theta)$ is the planar rotation, and $q_\parallel \ge q_\perp$ are the
along-track and cross-track noise intensities.

\subsection{Landmark Measurements}

A landmark $\ell$ provides a range-bearing measurement whose quality decays
smoothly with the distance $d_{i\ell} = \lVert x_i - \ell \rVert$ through the
logistic taper
\begin{equation}
  q_{i\ell} = \frac{1}{1 + \exp\!\big((d_{i\ell} - r_v)/w_v\big)},
  \label{eq:taper}
\end{equation}
which is close to one inside the visibility range $r_v$ and rolls off over a
width $w_v$, with no hard cutoff. Every landmark contributes information
weighted by its quality, and the posterior follows the information form of
the Kalman update for a direct position measurement, identical to the Joseph
form for $H = I$:
\begin{equation}
  \Sigma_i(s_k) = \Big( \Sigma_i^{-}(s_k)^{-1}
  + \sum_{\ell \in \mathcal{L}} q_{i\ell}\,\big(S_{i\ell} + \Sigma_\ell\big)^{-1} \Big)^{-1},
  \label{eq:lm}
\end{equation}
where $S_{i\ell} = R(\beta_{i\ell})\,\mathrm{diag}(\sigma_r^2, \sigma_b^2)\,R(\beta_{i\ell})^{\!\top}$
is the sensor noise rotated to the line-of-sight bearing $\beta_{i\ell}$.

\subsection{Inter-Agent Fusion}

Agents exchange position estimates at fixed arc-length intervals $\Lambda$.
Their estimates share history, so adding the information directly would
double count it, and we fuse with covariance intersection
\cite{julier1997cov}, which is consistent under unknown cross-correlation.
When agents $i$ and $j$ communicate at separation $d_{ij}$, agent $i$
receives $P_{j \to i} = (\Sigma_j + R_c)/w_{ij}$, where $R_c = \sigma_c^2 I$
is the communication noise and $w_{ij}$ is a taper of the form
\eqref{eq:taper} with half weight at the communication range $r_c$, and
updates
\begin{equation}
  \Sigma_i^{+} = \Big( \omega\,\Sigma_i^{-1} + (1-\omega)\,P_{j \to i}^{-1} \Big)^{-1},
  \label{eq:ci}
\end{equation}
with $\omega \in [0,1]$ chosen to minimize $\det\Sigma_i^{+}$. Agent $j$ is
updated in the same way, and as $w_{ij} \to 0$ the transferred information
vanishes.

\subsection{Optimization Problem}

The planning problem collects the objective and all constraints as
\begin{subequations}
\label{eq:opt}
\begin{align}
  & \min_{\{\theta_i(\cdot)\},\, S_1}\; S_1 \;\;\; \mathrm{s.t.}\;\;
    \det\big(\Sigma_1(S_1)\big)^{1/4} \le \bar{u}, \label{eq:term} \\
  & a_j^{\top} x_i(s) \ge b_j + r_a
      + z_\delta \sqrt{a_j^{\top}\big(\Sigma_i(s) + \Sigma_o\big)a_j},
      \label{eq:cc} \\
  & |\kappa_i(s)| \le 1/r_{\min}, \label{eq:kin} \\
  & x_i(0) = x_i^0,\;\; \Sigma_i(0) = \Sigma^0,\;\; x_1(S_1) = G.
      \label{eq:bc}
\end{align}
\end{subequations}

\textbf{Terminal uncertainty \eqref{eq:term}.} The primary must arrive with
bounded localization uncertainty. $\det(\Sigma)^{1/4}$ equals the standard
deviation $\sigma$ of an isotropic covariance $\sigma^2 I$.

\textbf{Obstacle avoidance \eqref{eq:cc}.} Every agent must clear every
obstacle with probability at least $1-\delta$ at every instant, so for each
agent $i$, obstacle $o$, and arc position $s$ the constraint must hold for at
least one face $j$ of $\mathcal{O}_o$. The margin covers the vehicle radius
$r_a$ and the positional uncertainty of both agent and obstacle, with
$z_\delta = \Phi^{-1}(1-\delta)$ and $\Phi$ the standard normal distribution
function \cite{blackmore2011chance}. A union bound over the $PM$
agent-obstacle pairs bounds the instantaneous probability of any collision by
$\Delta = PM\delta$.

\textbf{Kinematics \eqref{eq:kin}.} Each trajectory respects the minimum
turn radius $r_{\min}$, where $\kappa_i = \mathrm{d}\theta_i/\mathrm{d}s$ is
the curvature.

\textbf{Boundary conditions \eqref{eq:bc}.} All agents start from known
positions with a common prior covariance, and the primary must end at the
goal.

The belief dynamics do not appear as constraints: $\Sigma_i(s)$ is a
deterministic function of the decision variables through
\eqref{eq:dyn}--\eqref{eq:ci}, left implicit in \eqref{eq:opt}. The agents
are coupled only through \eqref{eq:ci}, so the supports enter the problem
only by shaping the primary's covariance, at a cost in primary length that grows as $\bar{u}$ tightens (Fig.~\ref{fig:ladder}).

\section{Method}
\label{sec:method}
CONSORT solves \eqref{eq:opt} in two stages, a joint A* search over a
heading-aware hexagonal lattice and a continuous B-spline refinement of the
resulting plan, summarized in Algorithm~\ref{alg:joint}.

\subsection{Heading-Aware Hexagonal Graph}

A corridor of hexagonal cells \cite{duszak2021hexgrid} of width
$w_{\mathrm{hex}}$ covers the line from start to goal. A graph node pairs a
cell with
one of six headings spaced $60^\circ$ apart, and three moves leave each
node: continue straight, or turn by $\pm 60^\circ$ and step into the cell
faced by the new heading. The heading changes by at most $60^\circ$ per
cell, which is the graph-level counterpart of \eqref{eq:kin}. Landmarks are
appended as sensor nodes without edges that enter the search only through
\eqref{eq:lm}, a goal node absorbs all six headings of the goal cell so that
the arrival heading is free, and obstacles act on beliefs during the search
rather than removing cells. All-pairs shortest-path distances are
precomputed on the graph and serve as the heuristic. 
With only six headings, a line aimed between two of them is traced by a zigzag
at most $1/\cos 30^\circ \approx 1.15$ times as long; refinement cuts these
corners, shortening the seed by up to 7\% (Sec.~\ref{sec:abl}).

\subsection{Joint Discrete Search}

The first stage runs an A*-style search \cite{hart1968astar} over the joint
state of the team. A search state stores, for every agent, its graph node,
its covariance, and the arc distance traveled. Each expansion advances
every agent by one edge, so the successors of a state are the Cartesian
product of the agents' three moves. The accumulated cost $g$ is the
primary's arc distance alone, and states are expanded in order of
$f = g + (1+\varepsilon)\,h$, where $h$ is the precomputed shortest-path
distance from the primary's node to the goal. Supports contribute no
heuristic cost, so $h$ is admissible, and $\varepsilon > 0$ gives a weighted
search \cite{pohl1970weighted} that trades a factor of at most
$(1+\varepsilon)$ on the seed length \cite{pearl1984heuristics} for fewer
expansions. Ties are broken toward lower primary
uncertainty and then toward fewer idle supports.

During expansion each agent's covariance is propagated along its edge by
\eqref{eq:prop}--\eqref{eq:lm}, and \eqref{eq:ci} is applied whenever two
agents reach a communication checkpoint at the same arc distance. A
successor is discarded if any agent violates \eqref{eq:cc} along its edge or
if a support travels farther than the primary. Pareto pruning keeps the
joint space manageable: at each joint node signature the search keeps a
frontier of $(g, \Sigma_1)$ labels and discards a new label if an existing
one has no larger $g$ and a covariance no larger in the positive
semidefinite order. The pruning is sound because the heuristic is shared at
the same node tuple and a larger covariance cannot become smaller under
further propagation.

\begin{algorithm}[t]
\SetInd{0.4em}{0.8em}
\DontPrintSemicolon
\small
\SetKwInOut{Input}{Input}\SetKwInOut{Output}{Output}
\Input{hex graph $\mathcal{G}$, start states $x_i^0$, $\Sigma^0$, goal $G$,
       bound $\bar{u}$, expansion budget $N$}
\Output{B-spline control points $x$ of every agent}
$h(\cdot) \leftarrow$ \textsc{AllPairsShortestPaths}$(\mathcal{G})$\;
$\mathit{open} \leftarrow \{\text{joint start}\}$;
  $\mathit{inc} \leftarrow \varnothing$; $\mathit{seed} \leftarrow \varnothing$\;
\While{$\mathit{open} \neq \varnothing$ and expansions $< N$}{
  pop the state minimizing $f = g + (1+\varepsilon)\,h$\;
  \If{primary at $G$ and \textnormal{\textsc{SeedSplineClear}}$(\mathrm{state})$}{
    \lIf{$\det(\Sigma_1)^{1/4} \le \bar{u}$}{$\mathit{seed} \leftarrow \mathrm{state}$; \textbf{break}}
    \lElseIf{$\det(\Sigma_1)^{1/4}$ lowest so far}{$\mathit{inc} \leftarrow \mathrm{state}$}
  }
  \ForEach{joint successor (every agent one edge)}{
    propagate $\Sigma_i$ by \eqref{eq:prop}--\eqref{eq:lm}; fuse by \eqref{eq:ci}\;
    discard if \eqref{eq:cc} fails or a support outruns the primary\;
    insert unless dominated in $(g, \Sigma_1)$ at its signature\;
  }
}
\lIf{$\mathit{seed} = \varnothing$}{$\mathit{seed} \leftarrow \mathit{inc}$, the least-violating candidate}
$x \leftarrow$ control points of the seed splines;
  $\mathit{feas} \leftarrow \min s(x) \ge 0$\;
$x^\star \leftarrow x$ if $\mathit{feas}$ else $\varnothing$;
  $x^{-} \leftarrow x$ \tcp*[r]{incumbents}
\ForEach{barrier stage; iterate until stationary or stalled}{
  $x' \leftarrow$ backtracked Adam step on $\nabla_x J(x;\mu)$\;
  \uIf(\tcp*[f]{recovery}){$\lnot\mathit{feas}$}{
    accept iff $J(x') < J(x)$; then $\mathit{feas} \leftarrow \min s(x') \ge 0$\;
  }
  \lElse{accept iff $J(x') < J(x)$ and $\min s(x') \ge 0$}
  update $x^\star$ (shortest feasible) and $x^{-}$ (least violating) with $x'$\;
  at stage end, decay $\mu$ only once $\mathit{feas}$\;
}
\Return $x^\star$ if $x^\star \neq \varnothing$ else $x^{-}$\;
\caption{CONSORT}
\label{alg:joint}
\end{algorithm}

The search is anytime with respect to \eqref{eq:term}. When a popped state
places the primary at the goal and the spline seed it induces
(Sec.~\ref{sec:cont}) clears all obstacles, the state is returned if
\eqref{eq:term} holds and otherwise kept as the incumbent if its terminal
uncertainty is the lowest seen so far. 
If the expansion budget is exhausted, the continuous stage attempts to recover feasibility for the incumbent, relaxing only the terminal bound, never obstacle clearance.

\subsection{Continuous B-Spline Refinement}
\label{sec:cont}

The seed is lifted into continuous space as clamped cubic B-splines
\cite{piegl1997nurbs}, one per agent, with the seed's cell centers as control points. Support paths are padded to the primary's node count, and
the primary's final control point is pinned to $G$. The decision vector $x$
stacks the free control points: the interior points of the primary and all
but the first point of each support, whose terminal position is free.

The constraints of \eqref{eq:opt} form a slack vector $s(x)$, feasible when
$s(x) \ge 0$: the terminal bound, evaluated by propagating
\eqref{eq:prop}--\eqref{eq:ci} along the sampled splines, the curvature
bound, sampled from the spline derivatives, the requirement that no support
spline be longer than the primary's, and obstacle clearance. For the latter,
each spline segment is enclosed in the convex hull of its MINVO control
points \cite{tordesillas2022minvo}, and each segment-obstacle pair is
committed to the face of \eqref{eq:cc} that the seed clears by the largest
margin, which fixes the homotopy class and replaces the disjunction over
faces with linear rows, one per hull vertex. The refinement then minimizes
the primary's length under the log-barrier merit function
\begin{equation}
  J(x;\mu) = S_1(x) + \sum\nolimits_k B\big(s_k(x);\mu\big),
  \label{eq:merit}
\end{equation}
\begin{equation}
  B(s;\mu) =
  \begin{cases}
    -\mu \log s, & s > \delta_b, \\[2pt]
    -\mu \big( \log\delta_b + \tfrac{s-\delta_b}{\delta_b} \big), & s \le \delta_b,
  \end{cases}
  \label{eq:barrier}
\end{equation}
where the tangent extension below the knee $\delta_b$ keeps $J$ finite for
violated constraints and supplies a constant restoring gradient, so one
objective serves both feasible and infeasible seeds alike.

We minimize $J$ with Adam \cite{kingma2015adam} on forward-difference
gradients and a backtracking line search over a schedule of barrier stages (Algorithm~\ref{alg:joint}): while a feasible trajectory has not been found, every descent step is accepted. Once one is found, only steps that remain feasible are accepted and $\mu$ decays. Finally, CONSORT returns the shortest feasible trajectory or the least violating if none were found.
\section{Results}
\label{sec:results}

\subsection{Experimental Setup}

We evaluate CONSORT on randomized corridor scenarios with one primary and
one support. 
Each scenario places the goal 600 to 900\,m from the start,
draws two to four landmarks with map covariances of 0.6 to 1.35\,m standard
deviation, placed 280 to 430\,m to the side of the direct line on
alternating sides so that every fix costs a detour, and draws one to three
convex obstacles of up to 220\,m across, centered on a triangular
distribution so that most scenarios have an obstacle on the direct line. The
lattice width is 150\,m, the minimum turn radius 40\,m, the visibility range
100\,m with a 15\,m taper, the communication interval 100\,m, and the
communication range 300\,m with a 20\,m taper. Dead-reckoning variance grows
at $q_\parallel = 0.25$\,m$^2$ per meter along track, a standard deviation
of 5\,m after 100\,m, and at one ninth of that across track. The obstacle
risk is $\delta = 0.05$ per agent-obstacle pair with a 5\,m vehicle radius,
the search runs with $\varepsilon = 0$ and $10^6$ expansions, and the
refinement runs 15 barrier stages sharing 1000 Adam iterations.

\begin{figure}[t]
  \centering
  \includegraphics[width=\columnwidth]{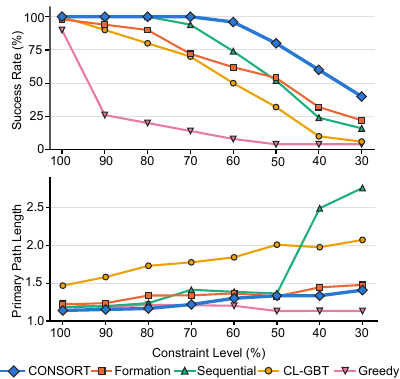}
  \caption{Baseline comparison over the 50 randomized scenarios as the terminal
    uncertainty bound tightens from 100\% to 30\% of the single-agent
    reference $U_\mathrm{ref}$. Top: fraction of scenarios solved. Bottom:
    mean primary path length over the solved scenarios, normalized by the
    single-agent reference length $L_\mathrm{ref}$ (standard deviations in
    Table~\ref{tab:sweep}).}
  \label{fig:length_success}
  \vspace{-0.3cm}
\end{figure}

For every scenario, we first plan a single unconstrained agent, which gives a
reference length $L_\mathrm{ref}$ and terminal uncertainty $U_\mathrm{ref}$.
The bound is then tightened from 100\% to 30\% of $U_\mathrm{ref}$ in steps
of 10\%, and every planner is run at each level. A planner that fails at one
level is counted as failing at the tighter ones. Scenarios in which a
straight line already meets the loosest bound pose no localization problem
and are discarded, which removed 34 of 84 draws to leave the 50 scenarios
reported here. Lengths are reported relative to $L_\mathrm{ref}$ over the
scenarios a planner solved, and wall-clock times are for the whole
process including start-up and graph construction. All planners run single-threaded in Julia 1.12 on a desktop with an AMD Ryzen 5 5600X six-core processor and 32\,GB of memory, one
scenario at a time.

\subsection{Baselines}

We compare against four planners adapted from adjacent literature and implemented in Julia. All share
the covariance model of Sec.~\ref{sec:pf}, the obstacle test, and the
refinement of Sec.~\ref{sec:cont}, so the comparison isolates how the team
is routed.

\textit{Greedy.} The primary commits to its shortest obstacle-clear route
without uncertainty reasoning, and at every step the support picks, among
its three successors, the move that most reduces the primary's uncertainty
at that step, the one-step information-gain rule of greedy exploration
\cite{bourgault2002info}. This removes lookahead from the support.

\textit{Formation.} The primary plans alone with a weighted A* over cells
and beliefs while the support holds a fixed slot in the primary's body
frame, one cell out at a $60^\circ$ multiple off its heading, as in
leader-referenced formation control \cite{balch1998formation}. Goal candidates
are accepted on the uncertainty the primary has with the support attached,
and both lateral slots are tried. This removes support routing.

\textit{Sequential.} Prioritized planning
\cite{erdmann1987multiple,silver2005cooperative}. The primary plans
first with the support parked at the start, under a bound slackened to
twice the threshold, the support then plans against the committed
primary to minimize the primary's terminal uncertainty, and the primary replans once against the finished support path. This avoids the joint
state space.

\textit{CL-GBT.} A cooperative-localization-guided belief tree, inspired by \cite{theurkauf2025coop}, who extend Gaussian
belief trees \cite{ho2022gbt} to grow one tree over the joint belief of a
team. We keep that structure, one rapidly exploring random tree over the
joint state in continuous space, with goal-biased sampling for the primary
and a cloning bias that places the support's sample on the primary's. Their joint Kalman belief with cross-correlations, their bound
on the probability that a planned inter-agent measurement falls out of
range, their robot-robot collision constraint and their re-branching step
are replaced by our belief model, which credits every scheduled fusion at a
range-tapered weight. Its iteration budget is set so that its compute
ceiling matches ours. This replaces exhaustive lattice search with sampling.

\subsection{Comparison With Baselines}

\begin{figure}[t]
  \centering
  \includegraphics[width=\columnwidth]{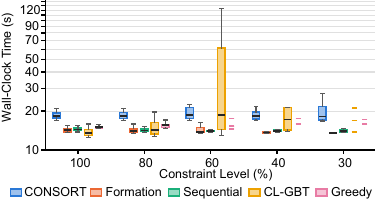}
  \caption{Wall-clock time per planner over successful runs at five constraint
    levels (log scale). Boxes span the interquartile range, whiskers extend
    to $1.5\times$ IQR, and runs beyond the whiskers are omitted. Levels with
    fewer than five solved scenarios are drawn as individual runs.}
  \label{fig:wall}
  \vspace{-1.5em}
\end{figure}

Fig.~\ref{fig:length_success} reports success rate and path length against
the constraint level, and Table~\ref{tab:sweep} lists the lengths with their
spread. Every planner but Greedy solves nearly every scenario at the loosest
level, and the planners separate as the bound tightens. CONSORT solves every
scenario down to 70\%, 96\% at 60\%, 80\% at 50\%, and 40\% at 30\%.
Sequential matches it down to 80\% but falls to 52\% at 50\% and 16\% at
30\%, and its paths grow to more than twice the reference length, since at a
tight bound the primary must detour to landmarks on its own before the
support is planned. Formation degrades more gracefully and is the strongest
baseline from 50\% down, with 54\% at half the reference bound and 22\% at
30\%. CL-GBT solves half the scenarios at 60\% and 6\% at 30\%, with paths
1.5 to 2 times the reference length because a random tree returns the first
path it finds. Greedy solves 90\% at the loosest level and almost nothing
below it, since a support without lookahead rarely reaches a landmark in
time.

The price of the higher success rate is modest. CONSORT's paths are among the shortest for the solved scenarios at every level, from $1.14\,L_\mathrm{ref}$ at
100\% to $1.41\,L_\mathrm{ref}$ at 30\%, with the smallest spread across
scenarios. Fig.~\ref{fig:wall} reports wall-clock time: a median of 18\,s
per run against 14\,s for Formation and Sequential and 16\,s for the
single-agent reference, most of it shared overhead. Only CL-GBT grows with
the bound, and its longest run took 730\,s.

\subsection{Ablation}
\label{sec:abl}

\begin{figure}[t]
  \centering
  \includegraphics[width=\columnwidth]{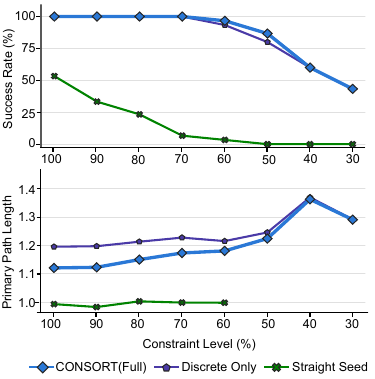}
  \caption{Ablation on a separate set of 30 scenarios drawn without the
    straight-line screen. Same axes as Fig.~\ref{fig:length_success}: top,
    success rate; bottom, mean path length over the solved scenarios
    (standard deviations in Table~\ref{tab:sweep}).}
  \label{fig:abl_length_success}
  \vspace{-0.3cm}
\end{figure}

\begin{figure}[t]
  \centering
  \includegraphics[width=0.987\columnwidth]{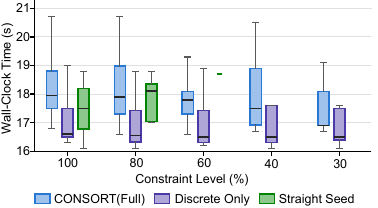}
  \caption{Wall-clock time of the ablation arms over successful runs, drawn as
    in Fig.~\ref{fig:wall} on a linear axis. The refinement adds about
    1.5\,s to the discrete stage. The straight seed's single run at 60\% is
    drawn as a tick.}
  \label{fig:abl_wall}
  \vspace{-0.3cm}
\end{figure}

The ablation asks what each stage contributes. It runs on a separate set of
30 scenarios drawn without the straight-line screen, since one arm is the
straight seed itself, and compares the full pipeline with the discrete stage
alone and with the refinement seeded from a straight line
(Fig.~\ref{fig:abl_length_success} and the bottom block of
Table~\ref{tab:sweep}). The discrete stage alone solves the same scenarios
down to 70\% and one or two fewer at 60\% and 50\%, with paths 2 to 7\%
longer at every level down to 50\%, which is the length the refinement
recovers by cutting the lattice corners. The straight seed solves 53\% of
the scenarios at the loosest level and none below 60\%. At the loosest level, 11 of its 16 solutions started with an obstacle violation of up to 55\,m that the refinement removed, and seven of these were up to 7\% shorter than the route of the full pipeline. The joint search finds the landmarks, and the
refinement makes the route short, at a cost of about 1.5\,s
(Fig.~\ref{fig:abl_wall}).

\section{Simulation Validation}
\label{sec:sim}
Every planner in Sec.~\ref{sec:results} propagates the same covariance
model, so the benchmark cannot tell whether that model is calibrated. We
therefore fly a planned trajectory with a vehicle model, a controller, and a
state estimator that share none of the planner's equations, and check
whether the certified constraints hold across noise seeds.

\subsection{Simulator and Vehicle}

\begin{figure}[tb]
  \centering\includegraphics[width=0.99\columnwidth]{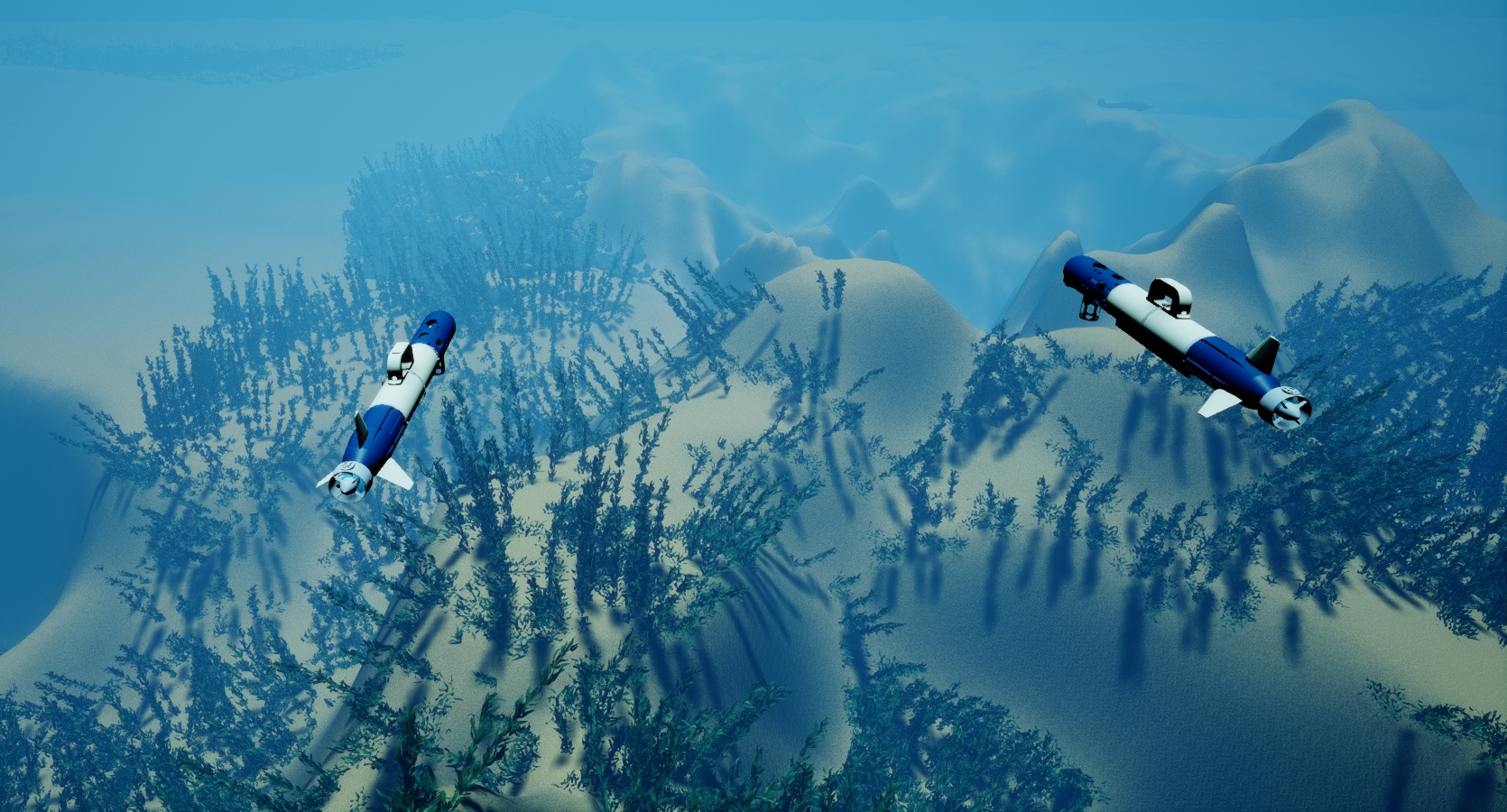}
  \caption{The two vehicles in HoloOcean~\cite{potokar2024holoocean}. Each is a
    torpedo-shaped autonomous underwater vehicle flying its planned spline at a
    fixed depth over the seabed of the simulator's open-water world.}
  \label{fig:holoocean}
  \vspace{-0.3cm}
\end{figure}

We use HoloOcean \cite{potokar2024holoocean} to simulate torpedo-shaped autonomous underwater vehicles with
six-degree-of-freedom Fossen dynamics \cite{fossen2021handbook}, driven by
the depth and heading autopilot of HoloOcean~2.0 \cite{romrell2025holoocean2}
at 30\,Hz at about 2.2\,m/s (Fig.~\ref{fig:holoocean}). An outer-loop waypoint follower
tracks the planned B-spline, resampled every 25\,m, with a 15\,m arrival
radius set by the turn radius, and it steers on the estimated position,
dead-reckoned forward between estimator updates, so estimation error feeds
back into the executed path. The planar field is placed on one depth plane
with the agents staggered in depth, so only agent-obstacle collisions are
under test. Three cadences are scheduled on arc length accumulated from
ground truth: the controller runs every tick, the estimator adds a pose node every 5\,m, and
an acoustic communication is attempted every 100\,m and succeeds only within
the planner's communication range. Measurements are ground truth with
additive Gaussian noise, so the simulation tests the consistency of the
belief model rather than the realism of any sensor.

\begin{table}[b]
  \centering
  \footnotesize
  \setlength{\tabcolsep}{4pt}
  \begin{tabular}{@{}lcccc@{}}
    \toprule
    & \multicolumn{4}{c}{Constraint level (\% of $U_\mathrm{ref}$)} \\
    \cmidrule(l){2-5}
    Method & 100 & 70 & 50 & 30 \\
    \midrule
    \multicolumn{5}{@{}l}{\emph{Baseline comparison, 50 screened scenarios}}\\
    CONSORT (ours)& {1.14$\pm$0.09} & {1.22$\pm$0.23} & 1.34$\pm$0.26 & {1.41$\pm$0.21} \\
    Formation     & 1.22$\pm$0.12 & 1.34$\pm$0.25 & {1.33$\pm$0.16} & 1.48$\pm$0.45 \\
    Sequential    & 1.18$\pm$0.14 & 1.41$\pm$0.92 & 1.36$\pm$0.50 & 2.76$\pm$1.51 \\
    CL-GBT        & 1.47$\pm$0.33 & 1.78$\pm$0.43 & 2.01$\pm$0.41 & 2.07$\pm$0.26$^\dagger$ \\
    Greedy        & 1.24$\pm$0.10 & 1.21$\pm$0.06 & 1.13$\pm$0.01$^\dagger$ & 1.13$\pm$0.01$^\dagger$ \\
    \addlinespace
    \multicolumn{5}{@{}l}{\emph{Ablation, 30 unscreened scenarios}}\\
    CONSORT (full) & 1.12$\pm$0.14 & {1.17$\pm$0.23} &{ 1.22$\pm$0.22} & {1.29$\pm$0.22} \\
    Discrete only          & 1.20$\pm$0.15 & 1.23$\pm$0.23 & 1.25$\pm$0.22 & {1.29$\pm$0.22} \\
    Straight seed + cont.\ & {0.99$\pm$0.04} & 1.00$\pm$0.00$^\dagger$ & -- & -- \\
    \bottomrule
  \end{tabular}
  \caption{Mean $\pm$ SD of primary path length over each method's own
    solved scenarios, normalized by $L_\mathrm{ref}$. Entries are not
    comparable across methods without the success rates in
    Figs.~\ref{fig:length_success} and~\ref{fig:abl_length_success}, since a
    method that solves only its easiest scenarios reports a short path.
    $^\dagger$ marks fewer than five solved scenarios. ``--''
    means none solved.}
  \label{tab:sweep}
\end{table}

\subsection{Factor-Graph State Estimator}

The estimator is an existing GTSAM factor-graph stack
\cite{kaess2012isam2,dellaert2017factor}, developed independently of the
planner; what we supply is its configuration and, through Table~
\ref{tab:noise}, its noise. It is an incremental smoother
\cite{kaess2012isam2} over $SE(3)$ pose nodes, one per 5\,m of arc for each
agent, returning the trajectory $\hat{\mathcal X}$ that minimizes the weighted
residuals $r_i$ of the factors $\mathcal F$ the graph holds,
\begin{equation}
  \hat{\mathcal X}=\arg\min_{\mathcal X}\;\sum_{i\in\mathcal F}
    \big\lVert r_i(\mathcal X)\big\rVert^{2}_{\Sigma_i}.
  \label{eq:fg}
\end{equation}

The graph is built in two layers. Each agent carries a dead-reckoning graph of
its own, holding a prior factor, a between-factor for the odometry of each node, a unary depth
factor, a unary three-axis orientation factor, and a range and bearing factor
to every landmark inside the visibility range. Landmarks are not free
variables: each is pinned by a prior carrying the planner's map covariance, so
the map is known and the problem is localization rather than mapping. This
graph is what the agent steers on, and for a support it is also the source of
the pose and the covariance its acoustic packets carry. The primary holds a
second graph, and the covariance compared against the planner's is read from
that one. It takes the primary's own nodes as they are created, and a
support's nodes in a batch when a communication succeeds, seeded with the pose
that support solved for itself, together with one inter-agent factor per
communication. Measurements enter both layers once rather than being drawn
again, and every one of them is ground truth plus additive Gaussian noise:
odometry is synthesized from the simulator's pose sensor, and there is no
inertial or Doppler sensor model.

A successful communication delivers the sender's most recent measured depth and orientation, and estimated pose and covariance. For robustness against dropout, we follow the origin state method of \cite{walls2015osm} and compute the relative pose and covariance between the pose most recently received and the one before it. The relative-pose factor is then inserted
into the primary's graph 
between the two vehicles, whose translation noise carries the planner's own
link weight.

\subsection{Monte Carlo Results}

\begin{table}[b]
      \centering
  \scriptsize
  \setlength{\tabcolsep}{3pt}
  \begin{tabular}{@{}llc@{}}
    \toprule
    Factor & Variables & $\sigma$ \\
    \midrule
    Odometry (between, per 5\,m node) & $\mathbf T_{k-1},\mathbf T_k$ & 1$^\circ$/axis; 0.22, 0.02, 0.01\,m \\
    Depth (unary) & $\mathbf T_k$ & 0.5\,m \\
    Compass (unary) & $\mathbf T_k$ & 0.86$^\circ$/axis \\
    Landmark az./el./range & $\mathbf T_k,\ell$ & 1.05$^\circ$, 1.05$^\circ$, 2.09\,m; $r<100$\,m \\
    Inter-agent relative pose & $\mathbf T^{s}_k,\mathbf T^{p}_k$ & 0.86$^\circ$/axis; $0.5/\sqrt{w}$\,m \\
    Initial pose prior & $\mathbf T_0$ & 10$^\circ$/axis; 1.15, 0.76, 0.5\,m \\
    \bottomrule
  \end{tabular}
  \caption{Factor noise of the simulated estimator, derived from the planner's model: odometry from the $q_\parallel$, the compass from $q_\perp$, landmark noise rescaled to one sighting per 5\,m node, communication noise $\sigma_c/\sqrt{w}$ for link weight $w$, and the start prior from $\Sigma^0$. The depth factor keeps the stack's 0.5\,m against the 0.01\,m injected.}
  \label{tab:noise}
\end{table}

\begin{figure}[t]
  \centering\includegraphics[width=0.97\columnwidth]{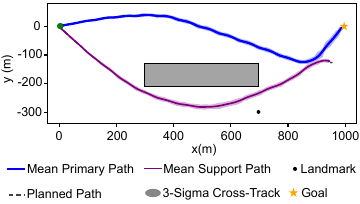}
  \caption{Closed-loop Monte Carlo of the behind-wall plan in HoloOcean over 30 runs with different noise seeds, showing the planned trajectories (dashed) against the mean executed trajectory of each agent with the $\pm3\sigma$
    cross-track envelope. The primary passes north of the wall while the
    support diverts south to the landmark and rejoins before the goal. No run penetrated the wall, and the planner's predicted terminal $\sigma$ of 1.80\,m compares with 1.70\,m measured from the true error.}
  \label{fig:mc_paths}
  \vspace{-0.3cm}
\end{figure}

We executed the two-agent behind-wall plan of Fig.~\ref{fig:mc_paths} 30 times with
different noise seeds. The scenario hides its one landmark behind a wall, so only the support can see it, under a terminal bound of
1.8\,m against which the planner predicts 1.80\,m. Every run reached the
goal within 2\% of the
planned length and none penetrated the wall, against an allocated collision
probability of $\Delta = 0.10$.
The estimator's terminal marginal was under the bound in 28 of 30 runs, with
the two exceptions at 1.85 and 1.96\,m. The sample covariance of the true
terminal error gives 1.70\,m, within 6\% of the prediction on the conservative side, while the
estimator's mean marginal of 1.51\,m is optimistic against both. The true
error also carried a mean offset of 2.6\,m along track and 0.8\,m across, a
bias no covariance check captures but one in the axis where neither the wall clearance nor the 15\,m arrival radius binds.
The baselines are not re-flown, as they share the belief model whose calibration is being tested.

\section{Discussion and Limitations}
The discrete stage is complete only with respect to the lattice. The lattice
is laid over the corridor without regard to the obstacles, so the lattice
paths that clear them are neither a subset nor a superset of the continuous
feasible set, and a route that exists only between two cells is not found. The straight-seed solutions of Sec.~\ref{sec:abl} show this.
The refinement can shorten a seed and recover a small violation but cannot
change its homotopy class. A finer lattice narrows the gap at the cost of a
larger graph, and the all-pairs heuristic is cubic in the number of nodes.

The belief model is Gaussian and assumes that the planned trajectory
receives the maximum-likelihood observation at each step
\cite{platt2010belief}. The simulation
tests calibration against an independent estimator fed ground truth plus Gaussian noise, so the Monte Carlo runs show that the planner's
covariance is consistent with that estimator under 
Table~\ref{tab:noise}'s noise, not that either matches a real acoustic modem or sonar. The along-track bias in the terminal error shows that a consistent
covariance does not rule out a biased estimate.

CONSORT is centralized, computed offline, and executed open loop on a
communication schedule fixed in arc length; replanning after a missed fix is
not addressed. The belief model credits every scheduled fusion at a
range-tapered weight, whereas \cite{theurkauf2025coop}
bound the probability that a planned inter-agent measurement is
unavailable. The joint search grows exponentially in the
number of agents, and all experiments use one support. In the benchmark it
cost 1.3 to 1.9 times the wall clock of the single-agent reference on
average, most of it fixed overhead, but the exponent will bind for larger
teams. Finally, the environment is planar with static convex obstacles of
known position covariance, and the vehicle is modeled by a turn-radius
bound alone.

\section{Conclusion}
We presented CONSORT, a centralized planner for a primary agent that must reach a
goal in minimum time in a GPS-denied environment while support agents are
routed to keep its localization uncertainty within a terminal bound.
Optimizing over mean trajectories and propagating covariance only for
constraint evaluation reduces the belief-space problem to a joint lattice
search followed by B-spline refinement, with chance-constrained obstacle
avoidance and a curvature bound enforced in both stages. On 50 randomized
scenarios CONSORT solved 80\% of the scenarios at half the reference
uncertainty against 54\% for the strongest baseline, at comparable path length. In Monte
Carlo trials with an independent factor-graph estimator, the executed trajectories remained collision-free and the planner's predicted terminal uncertainty fell within 6\% of the measured spread. Future work includes receding-horizon replanning,
scaling the joint search to larger teams, and hardware validation.

\bibliographystyle{IEEEtran}
\bibliography{refs}

\begin{thebibliography}{10}
\providecommand{\url}[1]{#1}
\csname url@samestyle\endcsname
\providecommand{\newblock}{\relax}
\providecommand{\bibinfo}[2]{#2}
\providecommand{\BIBentrySTDinterwordspacing}{\spaceskip=0pt\relax}
\providecommand{\BIBentryALTinterwordstretchfactor}{4}
\providecommand{\BIBentryALTinterwordspacing}{\spaceskip=\fontdimen2\font plus
\BIBentryALTinterwordstretchfactor\fontdimen3\font minus \fontdimen4\font\relax}
\providecommand{\BIBforeignlanguage}[2]{{%
\expandafter\ifx\csname l@#1\endcsname\relax
\typeout{** WARNING: IEEEtran.bst: No hyphenation pattern has been}%
\typeout{** loaded for the language `#1'. Using the pattern for}%
\typeout{** the default language instead.}%
\else
\language=\csname l@#1\endcsname
\fi
#2}}
\providecommand{\BIBdecl}{\relax}
\BIBdecl

\bibitem{roy_coastal_1999}
N.~Roy, W.~Burgard, D.~Fox, and S.~Thrun, ``\BIBforeignlanguage{en}{Coastal navigation-mobile robot navigation with uncertainty in dynamic environments},'' in \emph{\BIBforeignlanguage{en}{Proceedings 1999 {IEEE} {International} {Conference} on {Robotics} and {Automation} ({Cat}. {No}.{99CH36288C})}}, vol.~1.\hskip 1em plus 0.5em minus 0.4em\relax Detroit, MI, USA: IEEE, 1999, pp. 35--40.

\bibitem{roumeliotis2002dist}
S.~Roumeliotis and G.~Bekey, ``Distributed multirobot localization,'' \emph{IEEE Transactions on Robotics and Automation}, vol.~18, no.~5, pp. 781--795, 2002.

\bibitem{bahr2009coop}
A.~Bahr, J.~J. Leonard, and M.~F. Fallon, ``\BIBforeignlanguage{EN}{Cooperative {Localization} for {Autonomous} {Underwater} {Vehicles}},'' \emph{\BIBforeignlanguage{EN}{The International Journal of Robotics Research}}, vol.~28, no.~6, pp. 714--728, Jun. 2009.

\bibitem{theurkauf2025coop}
A.~Theurkauf, N.~Ahmed, and M.~Lahijanian, ``Multi-{Robot} {Motion} {Planning} with {Cooperative} {Localization},'' in \emph{2025 {IEEE}/{RSJ} {International} {Conference} on {Intelligent} {Robots} and {Systems} ({IROS})}, Oct. 2025, pp. 3962--3969.

\bibitem{potokar2024holoocean}
E.~Potokar, K.~Lay, K.~Norman, D.~Benham, S.~Ashford, R.~Peirce, T.~B. Neilsen, M.~Kaess, and J.~G. Mangelson, ``{HoloOcean}: A full-featured marine robotics simulator for perception and autonomy,'' \emph{IEEE Journal of Oceanic Engineering}, vol.~49, no.~4, pp. 1322--1336, Oct. 2024.

\bibitem{kaelbling1998pomdp}
L.~P. Kaelbling, M.~L. Littman, and A.~R. Cassandra, ``\BIBforeignlanguage{en}{Planning and acting in partially observable stochastic domains},'' \emph{\BIBforeignlanguage{en}{Artificial Intelligence}}, vol. 101, no. 1-2, pp. 99--134, May 1998.

\bibitem{papadimitriou1987mdp}
C.~H. Papadimitriou and J.~N. Tsitsiklis, ``The complexity of {Markov} decision processes,'' \emph{Mathematics of Operations Research}, vol.~12, no.~3, pp. 441--450, Aug. 1987.

\bibitem{prentice2009brm}
S.~Prentice and N.~Roy, ``\BIBforeignlanguage{EN}{The {Belief} {Roadmap}: {Efficient} {Planning} in {Belief} {Space} by {Factoring} the {Covariance}},'' \emph{\BIBforeignlanguage{EN}{The International Journal of Robotics Research}}, vol.~28, no. 11-12, pp. 1448--1465, Nov. 2009.

\bibitem{bry2011rrbt}
A.~Bry and N.~Roy, ``Rapidly-exploring {Random} {Belief} {Trees} for motion planning under uncertainty,'' in \emph{2011 {IEEE} {International} {Conference} on {Robotics} and {Automation}}, May 2011, pp. 723--730.

\bibitem{vandenberg2011lqgmp}
J.~van~den Berg, P.~Abbeel, and K.~Goldberg, ``\BIBforeignlanguage{EN}{{LQG}-{MP}: {Optimized} path planning for robots with motion uncertainty and imperfect state information},'' \emph{\BIBforeignlanguage{EN}{The International Journal of Robotics Research}}, vol.~30, no.~7, pp. 895--913, Jun. 2011.

\bibitem{platt2010belief}
R.~Platt~Jr., R.~Tedrake, L.~P. Kaelbling, and T.~Lozano-Pérez, ``\BIBforeignlanguage{en}{Belief space planning assuming maximum likelihood observations},'' in \emph{\BIBforeignlanguage{en}{Proceedings of Robotics: Science and Systems}}, Zaragoza, Spain, Jun. 2010.

\bibitem{paull2014auv}
L.~Paull, S.~Saeedi, M.~Seto, and H.~Li, ``{AUV} {Navigation} and {Localization}: {A} {Review},'' \emph{IEEE Journal of Oceanic Engineering}, vol.~39, no.~1, pp. 131--149, Jan. 2014.

\bibitem{julier1997cov}
S.~Julier and J.~Uhlmann, ``A non-divergent estimation algorithm in the presence of unknown correlations,'' in \emph{Proceedings of the 1997 {American} {Control} {Conference} ({Cat}. {No}.{97CH36041})}, vol.~4, Jun. 1997, pp. 2369--2373.

\bibitem{carrillo2013ci}
L.~C. Carrillo-Arce, E.~D. Nerurkar, J.~L. Gordillo, and S.~I. Roumeliotis, ``Decentralized multi-robot cooperative localization using covariance intersection,'' in \emph{2013 {IEEE}/{RSJ} {International} {Conference} on {Intelligent} {Robots} and {Systems}}, Nov. 2013, pp. 1412--1417.

\bibitem{liu2026risk}
H.~Liu, M.~Yu, Y.~Wang, W.~Zhang, and Q.~Tang, ``Explicit risk-{Aware} belief space planning for underwater active cooperative localization,'' \emph{Ocean Engineering}, vol. 346, p. 123710, Feb. 2026.

\bibitem{sharon2015cbs}
G.~Sharon, R.~Stern, A.~Felner, and N.~R. Sturtevant, ``Conflict-based search for optimal multi-agent pathfinding,'' \emph{Artificial Intelligence}, vol. 219, pp. 40--66, Feb. 2015.

\bibitem{erdmann1987multiple}
M.~Erdmann and T.~Lozano-Pérez, ``\BIBforeignlanguage{en}{On multiple moving objects},'' \emph{\BIBforeignlanguage{en}{Algorithmica}}, vol.~2, no. 1--4, pp. 477--521, Nov. 1987.

\bibitem{silver2005cooperative}
D.~Silver, ``\BIBforeignlanguage{en}{Cooperative {Pathfinding}},'' in \emph{\BIBforeignlanguage{en}{Proceedings of the AAAI Conference on Artificial Intelligence and Interactive Digital Entertainment (AIIDE)}}, vol.~1, no.~1, Jun. 2005, pp. 117--122.

\bibitem{duszak2021hexgrid}
P.~Duszak, B.~Siemiątkowska, and R.~Więckowski, ``\BIBforeignlanguage{en}{Hexagonal {Grid}-{Based} {Framework} for {Mobile} {Robot} {Navigation}},'' \emph{\BIBforeignlanguage{en}{Remote Sensing}}, vol.~13, no.~21, p. 4216, Oct. 2021.

\bibitem{feng2024bsrrt}
H.~Feng, Q.~Hu, Z.~Zhao, and X.~Feng, ``Smooth path planning under maximum curvature constraints for autonomous underwater vehicles based on rapidly-exploring random tree star with {B}-spline curves,'' \emph{Engineering Applications of Artificial Intelligence}, vol. 133, p. 108583, Jul. 2024.

\bibitem{blackmore2011chance}
L.~Blackmore, M.~Ono, and B.~C. Williams, ``Chance-{Constrained} {Optimal} {Path} {Planning} {With} {Obstacles},'' \emph{IEEE Transactions on Robotics}, vol.~27, no.~6, pp. 1080--1094, Dec. 2011.

\bibitem{tordesillas2022minvo}
J.~Tordesillas and J.~P. How, ``{MINVO} {Basis}: {Finding} {Simplexes} with {Minimum} {Volume} {Enclosing} {Polynomial} {Curves},'' \emph{Computer-Aided Design}, vol. 151, p. 103341, Oct. 2022.

\bibitem{hart1968astar}
P.~E. Hart, N.~J. Nilsson, and B.~Raphael, ``A formal basis for the heuristic determination of minimum cost paths,'' \emph{IEEE Transactions on Systems Science and Cybernetics}, vol.~4, no.~2, pp. 100--107, 1968.

\bibitem{pohl1970weighted}
I.~Pohl, ``Heuristic search viewed as path finding in a graph,'' \emph{Artificial Intelligence}, vol.~1, no. 3--4, pp. 193--204, Jan. 1970.

\bibitem{pearl1984heuristics}
J.~Pearl, \emph{Heuristics: Intelligent Search Strategies for Computer Problem Solving}.\hskip 1em plus 0.5em minus 0.4em\relax Reading, MA, USA: Addison-Wesley, 1984.

\bibitem{piegl1997nurbs}
L.~Piegl and W.~Tiller, \emph{The {NURBS} Book}, 2nd~ed.\hskip 1em plus 0.5em minus 0.4em\relax Springer, 1997.

\bibitem{kingma2015adam}
D.~P. Kingma and J.~Ba, ``Adam: {A} method for stochastic optimization,'' in \emph{International Conference on Learning Representations (ICLR)}, San Diego, CA, USA, 2015.

\bibitem{bourgault2002info}
F.~Bourgault, A.~A. Makarenko, S.~B. Williams, B.~Grocholsky, and H.~F. Durrant-Whyte, ``Information based adaptive robotic exploration,'' in \emph{IEEE/RSJ International Conference on Intelligent Robots and Systems}, vol.~1, 2002, pp. 540--545.

\bibitem{balch1998formation}
T.~Balch and R.~Arkin, ``Behavior-based formation control for multirobot teams,'' \emph{IEEE Transactions on Robotics and Automation}, vol.~14, no.~6, pp. 926--939, 1998.

\bibitem{ho2022gbt}
Q.~H. Ho, Z.~N. Sunberg, and M.~Lahijanian, ``Gaussian belief trees for chance constrained asymptotically optimal motion planning,'' in \emph{2022 International Conference on Robotics and Automation (ICRA)}, May 2022, pp. 11\,029--11\,035.

\bibitem{fossen2021handbook}
T.~I. Fossen, \emph{\BIBforeignlanguage{en}{Handbook of {Marine} {Craft} {Hydrodynamics} and {Motion} {Control}}}, 2nd~ed.\hskip 1em plus 0.5em minus 0.4em\relax John Wiley \& Sons, Apr. 2021.

\bibitem{romrell2025holoocean2}
B.~Romrell, A.~Austin, B.~Meyers, R.~Anderson, C.~Noh, and J.~G. Mangelson, ``A preview of {HoloOcean} 2.0,'' arXiv:2510.06160, Oct. 2025.

\bibitem{kaess2012isam2}
M.~Kaess, H.~Johannsson, R.~Roberts, V.~Ila, J.~J. Leonard, and F.~Dellaert, ``{iSAM2}: Incremental smoothing and mapping using the {B}ayes tree,'' \emph{The International Journal of Robotics Research}, vol.~31, no.~2, pp. 216--235, 2012.

\bibitem{dellaert2017factor}
F.~Dellaert and M.~Kaess, ``Factor {Graphs} for {Robot} {Perception},'' \emph{Foundations and Trends in Robotics}, vol.~6, no. 1-2, pp. 1--139, Aug. 2017.

\bibitem{walls2015osm}
J.~Walls, A.~Cunningham, and R.~Eustice, ``Cooperative localization by factor composition over a faulty low-bandwidth communication channel,'' \emph{Proceedings - IEEE International Conference on Robotics and Automation}, vol. 2015, pp. 401--408, 06 2015.

\end{thebibliography}

\end{document}